\documentclass[10pt,twocolumn,letterpaper]{article}

\usepackage{wacv}
\usepackage{times}
\usepackage{epsfig}
\usepackage{graphicx}
\usepackage{amsmath}
\usepackage{amssymb}
\usepackage{csquotes}

\usepackage{graphicx,subcaption}

\usepackage[breaklinks=true,bookmarks=false]{hyperref}

\usepackage{multirow}
\usepackage{array}
\newcolumntype{P}[1]{>{\centering\arraybackslash}p{#1}}
\newcolumntype{M}[1]{>{\centering\arraybackslash}m{#1}}

\DeclareMathOperator*{\argmax}{\mathrm{argmax}}

\wacvfinalcopy 

\def\wacvPaperID{425} 

\ifwacvfinal\fi
\title{Are Odds Really Odd? \\ Bypassing Statistical Detection of Adversarial Examples}

\author{Hossein Hosseini \qquad Sreeram Kannan \qquad Radha Poovendran\\
	University of Washington, Seattle, WA \\
}

\begin{document}

\maketitle

\begin{abstract}

Deep learning classifiers are known to be vulnerable to adversarial examples. 
A recent paper presented at ICML 2019 proposed a statistical test detection method based on the observation that logits of noisy adversarial examples are biased toward the true class. 
The method is evaluated on CIFAR-10 dataset and is shown to achieve $99\%$ true positive rate (TPR) at only $1\%$ false positive rate (FPR). 

In this paper, we first develop a classifier-based adaptation of the statistical test method and show that it improves the detection performance. 
We then propose Logit Mimicry Attack method to generate adversarial examples such that their logits {\it mimic} those of benign images.
We show that our attack bypasses both statistical test and classifier-based methods, reducing their TPR to less than $2.2\%$ and $1.6\%$, respectively, even at $5\%$ FPR. 
We finally show that a classifier-based detector that is trained with logits of mimicry adversarial examples can be evaded by an adaptive attacker that specifically targets the detector. Furthermore, even a detector that is iteratively trained to defend against adaptive attacker cannot be made robust, indicating that statistics of logits cannot be used to detect adversarial examples. 

\end{abstract}

\section{Introduction}\label{sec:intro}

Adversarial examples are inputs to machine learning models that an attacker intentionally designs to cause the model to make a mistake~\cite{Goodfellow2017openai}. 
One particular method for generating adversarial examples for image classifiers is adding small perturbations to benign inputs such that the modified image is misclassified by the model, but a human observer perceives the original content~\cite{biggio2013evasion,szegedy2013intriguing}. 

Several methods have been proposed for developing robust algorithms that classify adversarily perturbed examples into their ground-truth label, of which adversarial training is shown to be an effective approach~\cite{madry2017towards}. 
Several other defenses have been broken with adaptive iterative attacks~\cite{carlini2017towards,uesato2018adversarial,athalye2018obfuscated,athalye2018robustness,engstrom2018evaluating,carlini2019ami}. 
A different approach for defending against adversarial examples is to {\it detect} whether an input is adversarially manipulated. Several such detection methods have been proposed, but later shown to be ineffective~\cite{he2017adversarial,carlini2017adversarial,carlini2017magnet}.

\begin{figure}[t]
	\centering
	\includegraphics[width=1\linewidth]{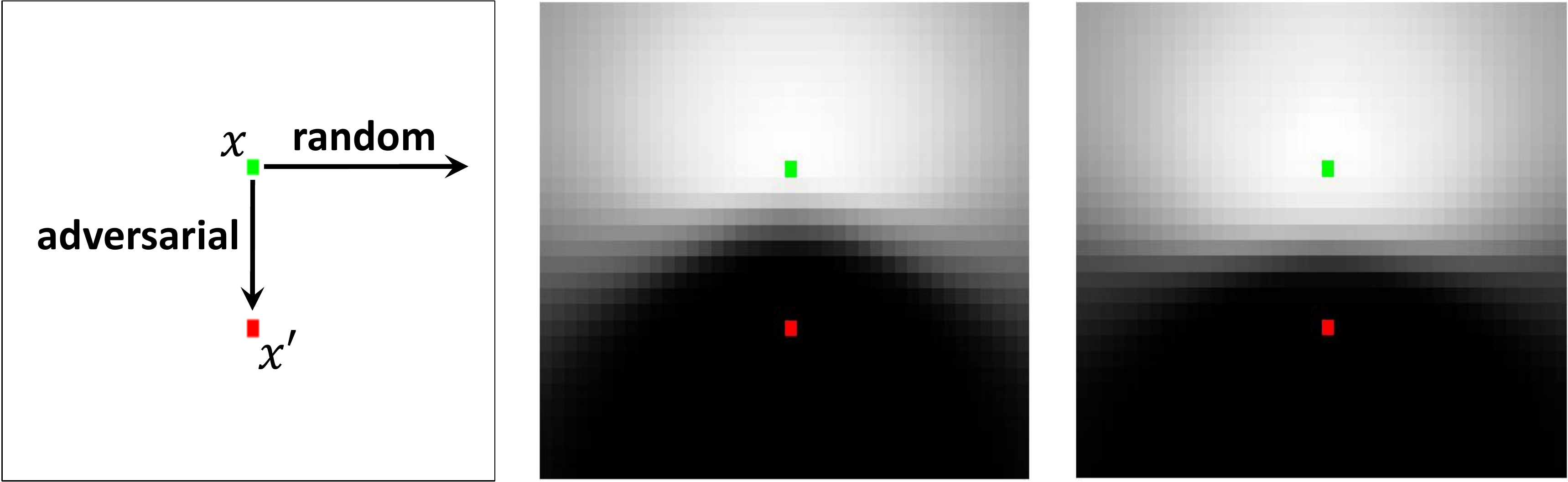}
	\caption{Average prediction probability of true class for samples across random and adversarial directions. 
		(Left) Illustration of directions and locations of benign sample, $x$, and adversarial example, $x'$. 
		(Middle) Probability map for adversarial examples generated using CW attack. 
		(Right) Probability map for adversarial examples generated using logit mimicry attack. 
		CW adversarial examples are embedded in a cone-like structure, referred to as {\it adversarial cone} in~\cite{roth2019odds}, indicating that adding noise increases expected probability of true class. 
		Mimicry adversarial examples, however, do not show such cone structure and are nearly as robust to noise as benign samples.}
	\label{fig:cone}
\end{figure}

A recent paper presented at ICML 2019 proposed a detection method based on a statistical test that measures how logits change under noise~\cite{roth2019odds}. The authors posited that robustness of logits are different, depending on whether the input is naturally generated or adversarially manipulated. Figure~\ref{fig:cone} visualizes this observation by measuring the average true probability across random and adversarial directions. The cone-like structure, referred to as {\it adversarial cone} in~\cite{roth2019odds}, indicates that the adversarial sample is ``surrounded'' by the true class. 
The authors then proposed to add noise to inputs as an approach to partially undo the effect of adversarial perturbations and, thus, detect and correctly classify adversarial examples. 
The method is evaluated on CIFAR-10 dataset and is shown to achieve $99\%$ true positive rate (TPR) at only $1\%$ false positive rate (FPR).

Unlike adversarial training method, several proposals for robust classifiers or detectors rely on ``subtle'' properties of models or benign or adversarial examples that are not ``tied'' to how the classification works, i.e., the model is not required to satisfy those properties to achieve good classification performance. 
Such defenses can be bypassed by {\it Mimicry Attack}, in which the attacker generates adversarial examples to not only fool the classifier, but also mimic the behavior of benign examples, where the specific behavior is derived from the mechanism that distinguishes between benign and adversarial examples. 
In~\cite{roth2019odds}, the authors observed that logits of adversarial examples are not robust to noise. In this paper, we show that this is a subtle property that is not fundamental to adversarial examples or deep learning classifiers, i.e., adversarial examples can be crafted such that they are misclassified by the model and their logits are indistinguishable from logits of benign examples. 
Our contributions are summarized in the following. 
\begin{itemize}	
	\item We first propose a classifier-based adaptation of the statistical test method. The proposed classifier takes logits of clean and noisy images as input and detects whether the image is adversarial. We show that such a classifier is able to identify anomalies in adversarial logits beyond their directionality toward the true class. Specifically, it improves the detection performance of the statistical test method against CW attack and succeeds even in small noise regimes where the test fails. 
	
	\item We then propose {\it Logit Mimicry Attack (LMA)}, 
	where the attacker collects logit profiles of clean and noisy benign images and crafts adversarial examples such that their logits are similar to those of benign images.
	We perform experiments on ResNet-56 network and CIFAR-10 dataset and show that our attack 
	bypasses both statistical test and classifier-based methods, reducing the TPR to less than $2.2\%$ and $1.6\%$, respectively, even at FPR of $5\%$. 
	Figure~\ref{fig:cone} visualizes the changes of average true probability across random and adversarial directions and shows that mimicry adversarial examples do not display the cone structure of the examples obtained from CW attack and are nearly as robust to noise as benign samples. 
	
	\item We then propose a mimicry detector, a classifier that is trained with logits of mimicry adversarial examples. Such a classifier improves the detection performance against LMA and achieves TPR of $62.7\%$ at $5\%$ FPR. We, however, show that an adaptive attacker that generates adversarial examples that evade both classifier and detector can successfully bypass the mimicry classifier, reducing the TPR to $1.7\%$ at  FPR of $5\%$. 
	
	\item We finally consider a detector that is iteratively trained against adaptive attack, i.e., at each iteration, adversarial examples are generated to bypass the detector and then the detector is trained to  classify them as adversarial. 
	Once training is done, the final detector is tested against the adaptive attack. 
	We show that such a detector only achieves the TPR of $2.9\%$ at $5\%$ FPR. The results indicate that logits of benign and adversarial exmaples are not fundamentally distinguishable, i.e., new adversarial examples can be always generated that evade both the network and any detector. 
\end{itemize}

The code for mimicry attack is available at \url{https://github.com/HosseinHosseini/Mimicry-Attack}.

\section{Preliminaries}\label{sec:Pre}

In this section, we provide the notations used in this paper and review the projected gradient descent method for generating adversarial examples.

\subsection{Notations}
Let $f: x\rightarrow z$ be a function that takes an image $x\in \mathbb{R}^d$, where $d$ is the number of pixels, and outputs the logit vector $z\in\mathbb{R}^k$, where $k$ is the number of classes. The probability vector $Y\in \mathbb{R}^k$ is then obtained as $Y=S(f(x))$, where $S$ is the softmax function, and the predicted label is $y=\argmax_i{Y_i}=\argmax_i{z_i}$. 
Let $\ell(x, y; \theta)$ denote the loss of the classifier with parameters $\theta$ on $(x, y)$. When holding $\theta$ fixed and viewing the loss as a function of $(x, y)$, we simply write $\ell(x, y)$. 

\subsection{Adversarial Examples}
We consider a class of adversarial examples for image classifiers where small (imperceptible) perturbations are added to images to force the model to misclassify them. 
We consider the case where the $L_{\infty}$ norm of perturbation is bounded. The attacker's problem is stated as follows:
\begin{align}\label{eq:attack} 
\mbox{given } x, &\mbox{ find } x' \\
\nonumber \mbox{s.t. } &\|x'-x\|_{\infty} \leq \epsilon_{\max} \hspace{0.2cm}\mbox{and}\hspace{0.2cm} f(x') \neq y,
\end{align}
where $x$ and $x'$ are the clean and adversarial samples, respectively, $y$ is the true label, and $\epsilon_{\max}$ is the maximum allowed absolute change to each pixel. 

Several optimization-based methods have been proposed for generating adversarial examples, including fast gradient sign method~\cite{goodfellow2014explaining}, iterative gradient method~\cite{kurakin2016adversarial}, and Carlini and Wagner (CW) attack~\cite{carlini2017towards}. It has been observed that the specific choice of optimizer is less important than choosing to use iterative optimization-based methods~\cite{madry2017towards}. 

We generate adversarial examples using the Projected Gradient Descent (PGD) method~\cite{kurakin2016adversarial,madry2017towards}, which provides a unified framework for iterative attacks independent of the specific optimization function. 
Let $n$ be the number of steps and $x^j$ be the image at step $j$. We have $x^0=x$ and $x'=x^n$. 
At step $j$, the image is updated as
\begin{align}\label{eq:pgd}
x^{j+1}=\Pi_{X+\mathcal{S}}(x^j + \epsilon_{\mathrm{step}} \hspace{0.05cm}\mathrm{sign}(v^j)),
\end{align}
where $v^j$ is the attack vector at step $j$, $\epsilon_{\mathrm{step}}$ is the added perturbation per step, and $\Pi_{x+\mathcal{S}}$ is the projection operator where $\mathcal{S}$ is the set of allowed perturbations. In the case of bounded $L_{\infty}$ constraint, projector clips each pixel within $\epsilon_{\max}$ of the corresponding pixel of original image $x$. 

The attack vector takes different forms depending on the attack goal. Generally, the attacker's goal is to maximize the loss on the defender's desired output or alternatively minimize the loss on attacker's desired output. 
These approaches lead to two common attacks respectively known as misclassification and targeted attacks, for which attack vectors are specified as follows:
\begin{itemize}
\item $v = \nabla_{x}\ell(x,y)$, for misclassification attack,
\item $v = -\nabla_{x}\ell(x,y_t)$, for targeted attack,
\end{itemize}
where $y_t\neq y$ is the attacker's desired target label.

\section{Detecting Adversarial Examples Based on Logit Anomalies}\label{sec:method}

In this section, we first review the logit-based statistical test proposed in~\cite{roth2019odds} for detecting adversarial examples and then describe our classifier-based adaptation of the method. 

\subsection{Statistical Test Proposed in~\cite{roth2019odds}}
In~\cite{roth2019odds}, a method is proposed for detecting adversarial examples and estimating their true class. The method is based on a statistical test that measures how logits change under noise. 
Let $f_{y_1,y_2}(x)=f_{y_2}(x)-f_{y_1}(x)$, where $f_{y}(x)$ denotes the logit of class $y$ for input $x$. 
Consider $f_{y_f,y}(x+\delta)$, where $y_f=\argmax_y f_y(x)$ is the predicted label and $\delta$ is the noise. 
Note that for $y\neq y_f$, $f_{y_f,y}(x)$ is negative and might increase with noise. 
The authors posited that common methods for generating adversarial examples result in perturbations that ``overfit to specifics of $x$ and are not robust,'' i.e., by adding noise to the adversarial example, the expected probability of the true class increases.

For an adversarial example $x$ and true label $y^*$, if the adversarial perturbation is not robust to noise, we will have $f_{y_f,y^*}(x+\delta) > f_{y_f,y^*}(x)$, meaning that adding noise will partially undo the effect of adversarial perturbation.
This is the basis of the statistical test, i.e., we obtain the logits of noisy inputs and check whether the expected logit value of a label increased under noise. Such an input is detected to be adversarially perturbed and the label with increasing logit is output as the correct class. 
The details of the method is provided in the following. 
\begin{itemize}	
	\item Let $g_{y_f,y}(x,\delta)=f_{y_f,y}(x+\delta)-f_{y_f,y}(x)$. The normalized term is obtained as  $\bar{g}_{y,z}(x,\delta)=(g_{y,z}(x,\delta)-\mu_{y,z})/\sigma_{y,z}$, where $\mu_{y,z}$ and $\sigma_{y,z}$ are mean and standard deviation of $g_{y,z}(x,\delta)$ computed on benign examples. 	
	Let $\bar{g}_{y_f,y}(x)=\mathbb{E}_{\delta}[\bar{g}_{y_f,y}(x,\delta)]$. 
	
	\item Inputs are flagged as adversarially manipulated if 
	\begin{align}
	\max_{y\neq y_f} \{\bar{g}_{y_f,y}(x)-\tau_{y_f,y}\}\geq 0,
	\end{align}
	where the thresholds $\tau_{y_f,y}$ are computed such that the true positive rate (TPR) is maximized, while the false positive rate (FPR) remains small (say $1\%$). If an input is detected to be an adversarial example, its correct label is predicted as $\argmax_{y\neq y_f} \{\bar{g}_{y_f,y}(x)-\tau_{y_y,y}\}$. 
\end{itemize}

\subsection{Our Classifier-Based Detection Method}\label{sec:classifier}
The detection method exploits the statistics of logits of clean and noisy inputs to distinguish between benign and adversarial samples. 
We note that instead of computing $\bar{g}_{y_f,y}(x)$ and the thresholds $\tau_{y_f,y}$, we can design a generic binary classifier that takes logits of clean and noisy images as input and detects whether the image is adversarial. The advantage of using such a classifier is that it can identify anomalies in adversarial logits beyond their directionality toward the true label and, hence, improves the detection performance. 
Our classifier-based detection method is described in the following. 
\begin{itemize}
	\item For validation data $x$, generate adversarial samples $x'$. 
	\item Compute logits of clean inputs and the average logits of noisy inputs, i.e., compute $f(x)$, $\bar{f}(x+\delta)=\mathbb{E}_{\delta}[f(x+\delta)]$, $f(x')$, $\bar{f}(x'+\delta)=\mathbb{E}_{\delta}[f(x'+\delta)]$. 
	\item Construct vectors $L(x)=f(x) || \bar{f}(x+\delta)$ and $L(x')=f(x') || \bar{f}(x'+\delta)$, where $||$ denotes concatenation. 
	\item Train a detector classifier with input-output pairs $(L(x),0)$ and $(L(x'),1)$. 
\end{itemize}

\begin{figure}[t]
	\centering
	\includegraphics[width=0.9\linewidth]{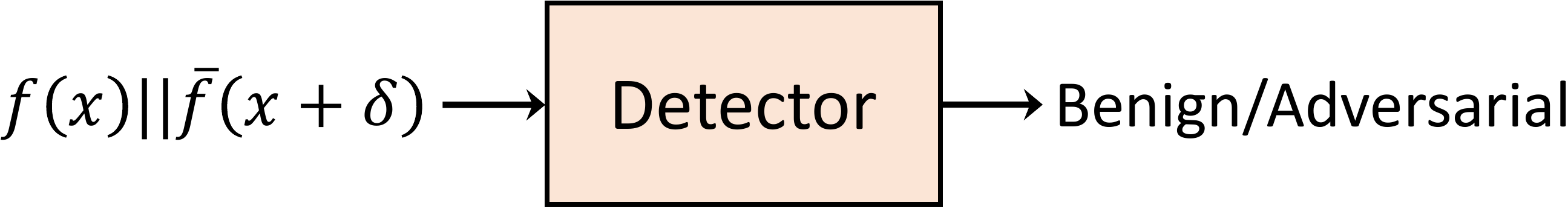}
	\caption{Classifier-based adaptation of the detection method proposed in~\cite{roth2019odds}. The classifier takes logits of clean image, $f(x)$, and the average logits of noisy images, $\bar{f}(x+\delta)=\mathbb{E}_{\delta}[f(x+\delta)]$, as input and detects whether the image is adversarial.}
	\label{fig:detector}
\end{figure}

In experiments, we use a neural network with two hidden layers of size $100$ and ReLU activation for the detector. 
Note that the statistical test method can be exactly implemented by such a network as well. Therefore, we expect our approach to be able to better distinguish between benign and adversarial examples. Figure~\ref{fig:detector} illustrates our classifier-based detection method. 

\section{Bypassing Logit-Based Detection Methods}

In this section, we first review the attack approach of computing adversarial perturbations on noisy images and then propose the logit mimicry attack for bypassing the logit-based detection methods. 

\subsection{Attacking with Noisy Inputs}
The statistical test method uses the property that adversarial examples are less robust to noise compared to benign examples. Based on this observation,~\cite{roth2019odds} suggested that, to bypass the defense, the attacker needs to craft perturbations that perform well in expectation under noise. In this attack, the PGD update is computed based on the following loss function:
\begin{align}
\hat{\ell}(x,y)=\alpha \ell(x,y) + (1-\alpha)\mathbb{E}_{\delta}[\ell(x+\delta,y)],
\end{align}
where $\alpha$ balances the attack updates for clean and noisy inputs. In experiments, we set $\alpha=0.25$ and compute the loss as an empirical average over $10$ noisy inputs, with the same noise source used for detection. 

\subsection{Logit Mimicry Attack}\label{sec:LMA}
Our classifier-based approach provides an additional insight into how the detection method can be bypassed. Recall that the detector takes the logits of clean images and the average logits of noisy images as input. Therefore, instead of computing perturbations that are robust to noise, the attacker can craft adversarial examples such that their logits {\it mimic} those of benign images. Specifically, an adversarial image that is classified as class $y$ must have similar logits to benign images that belong to the same class. Moreover, its expected logits of noisy examples must mimic average logits of noisy benign images that belong to that class. 
We call our attack {\it Logit Mimicry Attack}. 
The attack details are described in the following. 
\begin{itemize}
	\item Let $x$ be a validation image and $y_f$ be the predicted label, i.e., $y_f=\argmax_i f_i(x)$. For each class $y$, we compute the {\it logit profiles} of clean and noisy images respectively as follows:
	\begin{align}
	\begin{cases}
	h_y=\mathbb{E}_{x|y_f=y}[f(x)]\\
	h'_y=\mathbb{E}_{x|y_f=y}[\mathbb{E}_{\delta}[f(x+\delta)]].
	\end{cases}
	\end{align}
	
	\item Given a test image $x$, we aim to generate adversarial example $x'$ with the target class $y_t$. The attack losses on clean and noisy images are defined as follows:
	\begin{align}
	\begin{cases}
	\ell(x,y_t)=\|f(x)-h_{y_t}\|_2 ,\\
	\ell_{\delta}(x,y_t)=\|f(x+\delta)-h'_{y_t}\|_2.
	\end{cases}
	\end{align}
	
	\item The logit mimicry attack loss is then obtained as 
	\begin{align}\label{eq:LMA}
	\ell_{\tt LMA}(x,y_t) = \alpha\ell(x,y_t) + (1-\alpha)\ell_{\delta}(x,y_t),
	\end{align}
	where $\alpha$ balances the loss on clean and noisy inputs. In experiments, we set $\alpha=0.25$.
	
\end{itemize}

\bgroup
\def\arraystretch{1.2}
\begin{table*}[t]
	\centering
	\caption{True Positive Rate (TPR) of statistical test method proposed in~\cite{roth2019odds} and our classifier-based approach against different attacks and with False Positive Rates (FPR) of $1\%$ and $5\%$. Experiments are performed on ResNet-56 network and CIFAR-10 dataset. 
		Noisy images are generated as $x_n = \mbox{clip}(x+\epsilon v, -1,1)$, where $v\sim\mathcal{N}(0,1)$ and pixel range is in $[-1,1]$. 
		Adversarial examples are generated using targeted attack with maximum $L_{\infty}$ perturbation of $\epsilon=8/255\approx 3\%$ of the pixel dynamic range. 
		The results show that logit mimicry attack bypasses both detection methods. 
		(Higher numbers indicate better detection performance).
	}
	\begin{tabular}{|c|c|c|c|c|c|c|c|}
		\hline 
		\multirow{2}{*}{\bf Detection Method} & \multirow{2}{*}{\bf \large $\epsilon$} & \multicolumn{2}{c|}{\bf CW} & \multicolumn{2}{c|}{\bf CW with Noisy Inputs} & \multicolumn{2}{c|}{\bf Logit Mimicry}\\
		\cline{3-8}
		&&{\bf FPR$=1\%$} & {\bf FPR$=5\%$} &{\bf FPR$=1\%$} & {\bf FPR$=5\%$} &{\bf FPR$=1\%$} & {\bf FPR$=5\%$} \\
		\hline
		\hline
		\multirow{4}{*}{\bf Statistical Test} 
		& $0.01$ & $38.2\%$ & $55.4\%$ & $29.0\%$ & $51.8\%$ & $0.1\%$ & $0.9\%$\\
		\cline{2-8}
		& $0.1$ & $96.6\%$ & $99.9\%$ & $3.6\%$ & $48.8\%$ & $0.3\%$ & $2.2\%$\\
		\cline{2-8}
		& $1$ & $96.1\%$ & $98.4\%$ & $58.4\%$ & $69.2\%$ & $0.0\%$ & $0.0\%$\\
		\cline{2-8}
		& $10$ & $96.9\%$ & $98.6\%$ & $77.6\%$ & $88.2\%$ & $0.0\%$ & $0.0\%$\\
		\hline
		\hline
		\multirow{4}{*}{\bf Classifier-based} 
		& $0.01$ & $97.8\%$ & $99.4\%$ & $99.2\%$ & $99.6\%$ & $0.0\%$ & $0.0\%$\\
		\cline{2-8}
		& $0.1$ & $99.9\%$ & $100\%$ & $74.9\%$ & $93.5\%$ & $0.0\%$ & $1.6\%$\\
		\cline{2-8}
		& $1$ & $99.5\%$ & $99.7\%$ & $69.8\%$ & $81.6\%$ & $0.0\%$ & $0.2\%$\\
		\cline{2-8}
		& $10$ & $98.1\%$ & $99.0\%$ & $83.9\%$ & $92.0\%$ & $0.0\%$ & $0.1\%$\\
		\hline
	\end{tabular}
	\label{table:res}
\end{table*}
\egroup

\section{Experimental Results}\label{sec:attack}

In this section, we present the experimental results of evaluating statistical test and classifier-based methods against different attacks. 

\subsection{Attack Setup}
The experiments are performed on ResNet-56 network~\cite{he2016identity} and CIFAR-10 dataset~\cite{krizhevsky2009learning}. The network achieves $92.7\%$ accuracy on CIFAR-10 test data.
As suggested in~\cite{roth2019odds}, for each image, we generate $256$ noisy images as $x_n = \mbox{clip}(x+\epsilon v, -1,1)$, where $v\sim\mathcal{N}(0,1)$ 
and $\epsilon\in\{10^{-2}, 10^{-1}, 1, 10\}$. Note that the pixel range is in $[-1,1]$. 
We train the detector with $1000$ images of the CIFAR-10 validation set, from which $800$ images are used for training and the rest are used for validation. 
We then test the detector on $1000$ test images of CIFAR-10 dataset. 

The results are presented for targeted attack with PGD method. 
We generate adversarial examples using $50$ steps of PGD with $L_{\infty}$ perturbation of $\epsilon_{\max}=8/255\approx3\%$ of the pixel dynamic range, as done in~\cite{roth2019odds}. 
In all cases, we set the attack hyperparameters such that the attack success rate in evading the classifier is more than $99\%$. 
Unless stated otherwise, we use the CW loss function~\cite{carlini2017towards} in our attacks, which maximizes the logit of the targeted class, while minimizing the largest value of other logits as
\begin{align}
\ell(x,y_t)=(-f_{y_t}(x)+\max_{y\neq y_t}f_y(x)+\kappa)_+,
\end{align}
where higher values of $\kappa$ generates adversarial examples with higher confidence. We set $\kappa=50$ in our experiments. 
We use CW's approach, since it computes the loss using the logits and, hence, is more relevant to the detection method proposed in~\cite{roth2019odds}. 

\vspace{0.1cm}
\noindent{\bf Threat Model.} 
Various threat models have been considered for defending against evasion attacks. Commonly used threat models are white-box (complete knowledge of the system) and black-box (knowledge of training procedure and architecture of target model, but no access to the trained model itself). 
As stated in~\cite{carlini2019evaluating}, ``the guiding principle of a defense's threat model is to assume that the adversary has complete knowledge of the inner workings of the defense.'' That is, even in black-box setting, it must be assumed that the attacker is fully aware of the defense mechanism. 
In the field of security, this guideline is derived from Kerckhoffs's principle~\cite{kahn1996codebreakers}, which is stated also as ``The enemy knows the system'' by Claude Shannon~\cite{shannon1949communication}. 

In~\cite{roth2019odds}, the authors claimed robustness in white-box setting. However, most of the analysis and experiments were performed for the case where the attacker is not aware of the defense. 
The authors only briefly discussed attacking with noisy inputs as a defense-aware attack, which we show is not the right countermeasure against the defense. 
In our experiments, we consider white-box threat model and assume that the attacker is fully aware of the defense method.

\subsection{Attack Results}

\begin{figure*}[t]
	\centering
	\begin{subfigure}[t]{0.33\textwidth}
		\centering
		\includegraphics[width=0.95\linewidth]{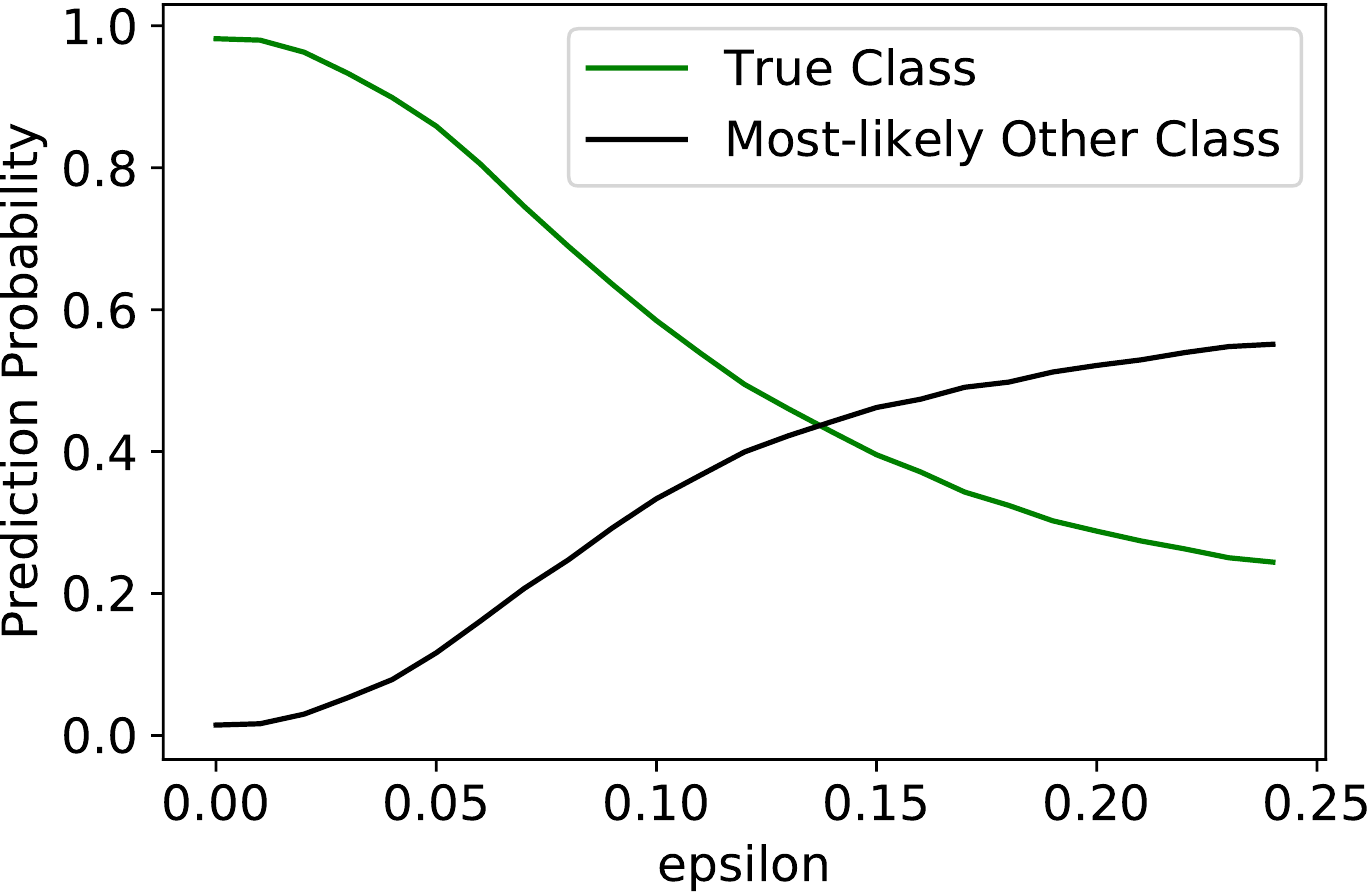}
		\caption{\small Benign examples.}
	\end{subfigure}\hspace{0.0cm}
	\begin{subfigure}[t]{0.33\textwidth}
		\centering
		\includegraphics[width=0.95\linewidth]{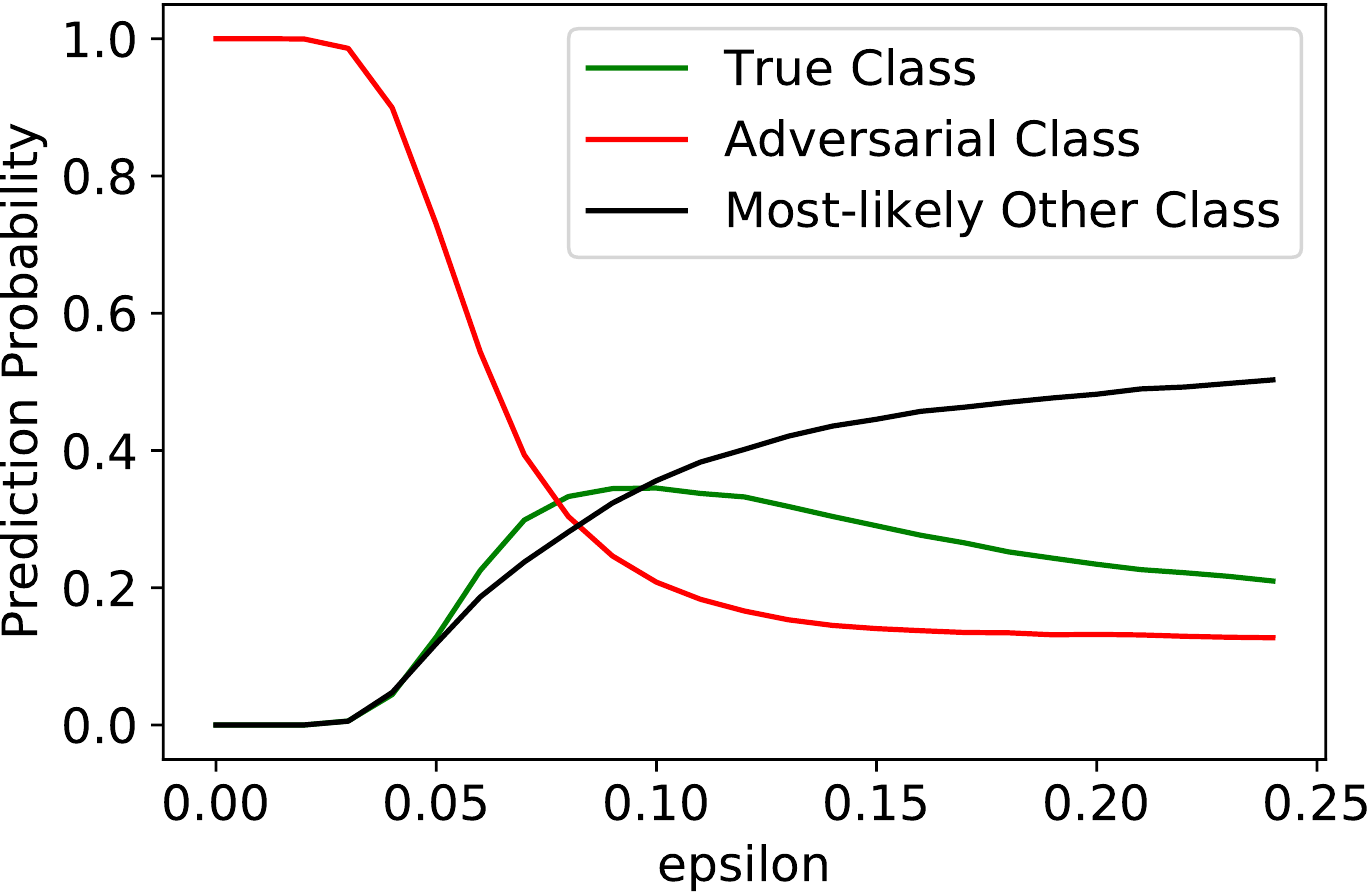}
		\caption{\small CW adversarial examples.}
	\end{subfigure}\hspace{0.0cm}
	\begin{subfigure}[t]{0.33\textwidth}
		\centering
		\includegraphics[width=0.95\linewidth]{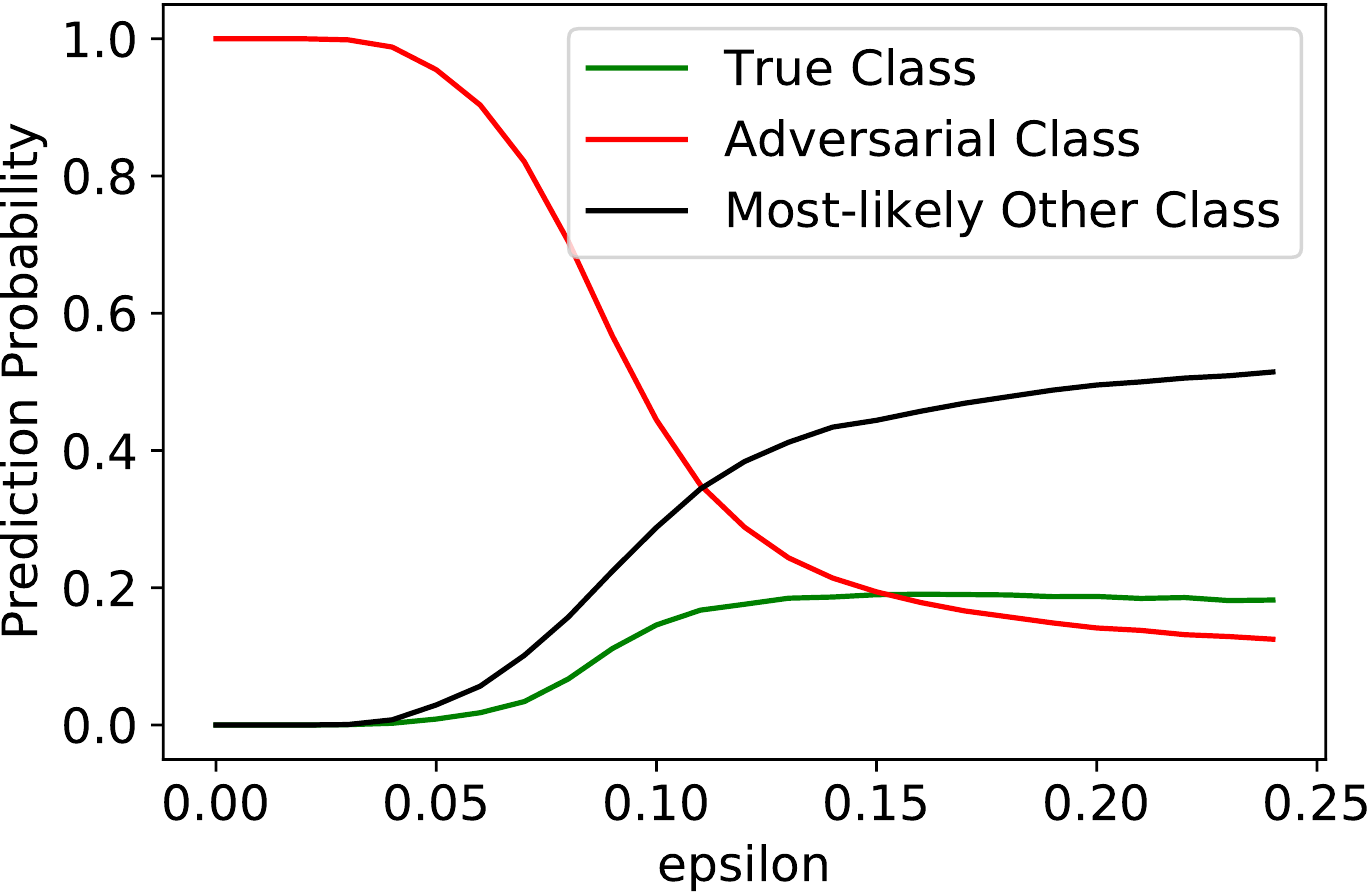}
		\caption{\small Mimicry adversarial examples generated with $\epsilon=0.1$.}
	\end{subfigure}\hspace{0.0cm}
	\caption{Average prediction probability for noisy inputs, $x_n=\mbox{clip}(x+\epsilon v, -1,1)$, where $v\sim\mathcal{N}(0,1)$. 
		In CW adversarial examples, there is a range of $\epsilon$, in which the probability of predicted (adversarial) class is less than the true class. This observation implies that noisy adversarial examples are biased toward the true class. In mimicry adversarial examples, however, the true class is never the most likely class.}
	\label{fig:curves}
\end{figure*}

\begin{figure}[t]
	\centering
	\begin{subfigure}[t]{0.1\textwidth}
		\centering
		\includegraphics[width=1\linewidth]{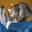}
	\end{subfigure}\hspace{0.0cm}
	\begin{subfigure}[t]{0.1\textwidth}
		\centering
		\includegraphics[width=1\linewidth]{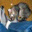}
	\end{subfigure}\hspace{0.0cm}
	\begin{subfigure}[t]{0.1\textwidth}
		\centering
		\includegraphics[width=1\linewidth]{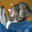}
	\end{subfigure}\\\vspace{0.1cm}
	\begin{subfigure}[t]{0.1\textwidth}
		\centering
		\includegraphics[width=1\linewidth]{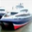}
	\end{subfigure}\hspace{0.cm}
	\begin{subfigure}[t]{0.1\textwidth}
		\centering
		\includegraphics[width=1\linewidth]{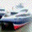}
	\end{subfigure}\hspace{0.cm}
	\begin{subfigure}[t]{0.1\textwidth}
		\centering
		\includegraphics[width=1\linewidth]{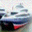}
	\end{subfigure}
	\caption{Samples of original and adversarial examples of CIFAR-10 dataset. {(Left)} original images, (Middle) adversarial examples generated using CW attack, (Right) adversarial examples generated using logit mimicry attack.}
	\label{fig:samples}
\end{figure}

\noindent{\bf CW Attack.} 
Table~\ref{table:res} shows the attack results for different settings. 
As can be seen, when noise is large enough, the statistical test method achieves more than $96\%$ true positive rate (TPR) at $1\%$ and $5\%$ false positive rates (FPR). The classifier-based approach achieves higher TPR especially when the noise is too small, indicating that it can better identify anomalies in adversarial logits. 
The results verify the observation that adversarial examples that are generated by only maximizing the loss on the true class are less robust to noise compared to benign examples. 
Figure~\ref{fig:curves} visualizes this property by measuring the average prediction probabilities of different classes versus noise. It can be seen that, in CW adversarial examples, there is a range of $\epsilon$, in which the true class is the most likely class, indicating that noisy CW adversarial examples are, indeed, biased toward the true class. 


\vspace{0.1cm}
\noindent{\bf CW Attack with Noisy Inputs.} 
As can be seen in Table~\ref{table:res}, adding noise to inputs at each PGD step reduces the detection performance of the statistical test method, but the method still achieves high TPR. Also, the attack approach is less effective against the classifier-based method, suggesting that logits of adversarial examples generated by noisy CW attack contain anomalies that can be identified by a more complex detector. 

\vspace{0.1cm}
\noindent{\bf Logit Mimicry Attack.} 
The mimicry attack bypasses both statistical test and classifier-based methods, resulting in less than $2.2\%$ TPR in all cases. The results show that such statistics-based defense mechanisms could be evaded by crafting adversarial examples that mimic the behavior of benign samples. 
Figure~\ref{fig:curves} shows the behavior of prediction probabilities versus noise and confirms that, unlike CW attack, in mimicry adversarial examples, the true class is never the most likely class, i.e., the directionality toward the true class has been eliminated. 
Figure~\ref{fig:samples} also shows samples of adversarial examples generated using CW and mimicry attacks. As can be seen, both attacks could evade the classifier by adding imperceptible perturbations to images.

\section{Does There Exist Any Logit-Based Robust Detector?}
In previous section, we showed that a classifier-based detector that is trained with CW adversarial examples can be bypassed with logit mimicry attack. 
The detector can be similarly trained with adversarial examples obtained using mimicry attack. In this section, we investigate whether such a classifier can detect mimicry adversarial examples and how it can be attacked. 

\vspace{0.1cm}
\noindent{\bf Training Detector with Mimicry Adversarial Examples.}
We follow the same approach of training a binary classifier proposed in Section~\ref{sec:classifier}, but generate adversarial examples using the logit mimicry attack. The attack results are provided in Table~\ref{table:adaptive}. As can be seen, the detector achieves $62.7\%$ TPR at $5\%$ FPR, indicating that the logits of mimicry adversarial examples are highly distinguishable from those of benign samples. 

\vspace{0.1cm}
\noindent{\bf Adaptive Attacker.}
To evade the adaptive detector, the attacker must craft adversarial examples that fool both the main network  and the detector.
Specifically, adversarial examples must be generated by minimizing the network's loss on targeted class and the detector's loss on benign class. We use logit mimicry attack to evade the network, since it also helps in bypassing the detector. 
The attack method is described in the following. 
\begin{itemize}
	\item Let $f$ be a function that takes a sample $x$ as input and outputs the logits. Let also $D$ be a binary classifier that takes logits of the clean sample and the average logits of noisy samples, and outputs zero if the image is benign and one otherwise. Let $p_{\tt benign}$ be the prediction probability of the benign class. We define the detector's loss on sample $x$ as $\ell_{D}(x) = -\log(p_{\tt benign})$.
	
	\item The overall attacker's loss is then computed as
	\begin{align}
	\ell_{\tt Adaptive}(x,y_t) = \alpha\ell_{\tt LMA}(x,y_t) + (1-\alpha)\ell_D(x),
	\end{align}
	where $\ell_{\tt LMA}(x,y_t)$ is defined in Equ.~\ref{eq:LMA}. In experiments, we set $\alpha=0.25$. 
\end{itemize} 
The adaptive attack achieves $100\%$ success rate in evading the network and, as can be seen in Table~\ref{table:adaptive}, reduces the TPR of the adaptive detector to $1.7\%$. 

\bgroup
\def\arraystretch{1.1}
\begin{table}[t]
	\centering
	\caption{True Positive Rate (TPR) of classifier-based detection methods. 
		Experiments are performed on ResNet-56 network and CIFAR-10 dataset. 
		Noisy images are generated as $x_n = \mbox{clip}(x+\epsilon v, -1,1)$, where $v\sim\mathcal{N}(0,1)$ and $\epsilon=0.1$ (pixel range is in $[-1,1]$). 
		Adversarial examples are generated using targeted attack with maximum $L_{\infty}$ perturbation of $\epsilon=8/255\approx 3\%$ of the pixel dynamic range. 
		The results show that any classifier-based detection method fails against adaptive attacker. 
		(Higher numbers indicate better detection performance).}
	\begin{tabular}{|p{2cm}|c|c|c|c|}
		\hline 
		\multirow{2}{*}{\bf \scriptsize Detection Method} & \multicolumn{2}{c|}{\bf \scriptsize Logit Mimicry Attack} & \multicolumn{2}{c|}{\bf \scriptsize Adaptive Attack}\\
		\cline{2-5}
		&{\bf \tiny FPR$=1\%$} & {\bf \tiny FPR$=5\%$} &{\bf \tiny FPR$=1\%$} & {\bf \tiny FPR$=5\%$} \\
		\hline
		\hline
		{\bf \scriptsize Adaptive Detector} 
		& {\small $26.6\%$} & {\small $62.7\%$} & {\small $0.4\%$} & {\small $1.7\%$}\\
		\hline
		{\bf \scriptsize Iteratively-trained Detector} 
		& {\small $49.4\%$} & {\small $83.1\%$} & {\small $0.6\%$} & {\small $2.9\%$}\\
		\hline
	\end{tabular}
	\label{table:adaptive}
\end{table}
\egroup

\begin{figure}[t]
	\centering
	\includegraphics[width=1\linewidth]{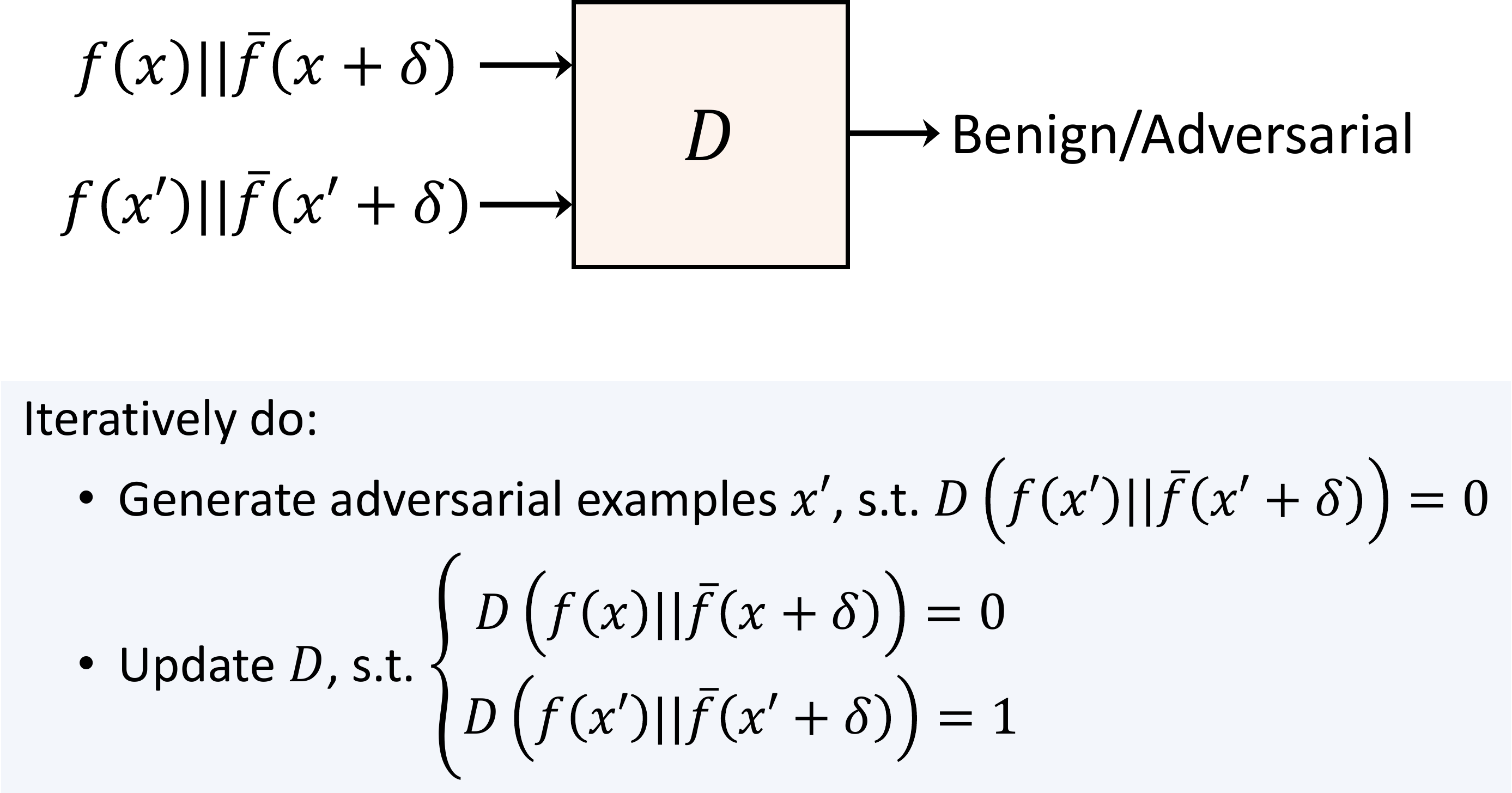}
	\caption{Iteratively training the detector. Detector is a classifier that takes logits of clean image, $f(x)$, and the average logits of noisy images, $\bar{f}(x+\delta)=\mathbb{E}_{\delta}[f(x+\delta)]$, as input and detects whether the image is adversarial. We iteratively run adaptive attack against the detector and fine-tune the detector with the new adversarial examples.}
	\label{fig:adaptive}
\end{figure}

\vspace{0.1cm}
\noindent{\bf Iteratively-Trained Detector, Adaptive Attacker.} 
We showed that adaptive attack successfully bypasses the adaptive detector. The question is does there exist any logit-based detector that is robust against adaptive attack?
To answer this question, we iteratively update the detector against the adaptive attack, i.e., we train a detector and then iteratively run the adaptive attack against it and fine-tune the detector with the new adversarial examples. Once training is done, we run adaptive attack on final detector. 
In experiments, we train the detector for $25$ iterations. At each iteration, the detector is trained for five epochs with $1000$ adaptive adversarial examples as well as $1000$ benign samples. Figure~\ref{fig:adaptive} illustrates the training procedure of iteratively-trained detector.

As can be seen in Table~\ref{table:adaptive}, while such a detector achieves high TPR against the logit mimicry attack, it completely fails against the adaptive attacker. That is, even after iteratively training the classifier to detect adaptive adversarial examples, we can always find new examples that evade both network and detector. 
The results indicate that distinguishability of benign and adversarial logits is not a fundamental property of deep convolutional networks and, hence, cannot be used to detect adversarial examples.

\section{Conclusion}\label{sec:conc}

A recent paper presented at ICML 2019 proposed a statistical test method for detecting adversarial examples. 
The method is designed based on the observation that logits of noisy adversarial examples ``tend to have a characteristic direction'' toward the true class, whereas it does not have any specific direction if the input is benign. 
In this paper, we first developed a classifier-based adaptation of the statistical test method, which improves the detection performance. 
We then proposed the Logit Mimicry Attack (LMA) method to generate adversarial examples such that the logits of clean and noisy images mimic the distribution of those of benign samples. 
We showed that LMA successfully bypasses both the statistical test and classifier-based methods. 
We finally evaluated the robustness of a detector that is iteratively trained to detect adversarial examples that are specifically generated to bypass it. We showed that even such a detector fails against the adaptive attack, indicating that adversarial logits can be made to mimic any characteristic behavior of benign logits.

%

{\small
\bibliographystyle{ieeetr}
\bibliography{Main}
}

\end{document}